\documentclass[sigconf]{aamas}  

\usepackage{booktabs}
\usepackage[utf8]{inputenc} 
\usepackage[T1]{fontenc}    
\usepackage{url}            
\usepackage{booktabs}       
\usepackage{amsfonts}       
\usepackage{nicefrac}       
\usepackage{microtype}      
\usepackage{caption}

\usepackage{url}
\usepackage{algorithm}
\usepackage{algorithmic}
\usepackage{graphicx}
\usepackage{amsthm}
\usepackage{amssymb}
\usepackage{multicol}
\usepackage{amsmath}

\usepackage{flushend}
\setcopyright{ifaamas}  
\copyrightyear{2020} 
\acmYear{2020} 
\acmDOI{} 
\acmPrice{} 
\acmISBN{} 
\acmConference[AAMAS'20]{Proc.\@ of the 19th International Conference on Autonomous Agents and Multiagent Systems (AAMAS 2020)}{May 9--13, 2020}{Auckland, New Zealand}{B.~An, N.~Yorke-Smith, A.~El~Fallah~Seghrouchni, G.~Sukthankar (eds.)}  

\begin{document}

\title{A Story of Two Streams: Reinforcement Learning Models \\ from Human Behavior and Neuropsychiatry}  



%

\author{Baihan Lin}
\affiliation{%
 \institution{Columbia University}
 \city{New York} 
 \state{NY} 
 \country{USA}
 \postcode{10027}
}
\email{baihan.lin@columbia.edu}

\author{Guillermo Cecchi}
\affiliation{%
 \institution{IBM Research}
 \city{Yorktown Heights} 
 \state{NY}
 \country{USA}
 \postcode{10598}}
\email{gcecchi@us.ibm.com}

\author{Djallel Bouneffouf}
\affiliation{%
 \institution{IBM Research}
 \city{Yorktown Heights} 
 \state{NY}
 \country{USA}
 \postcode{10598}}
\email{djallel.bouneffouf@ibm.com}

\author{Jenna Reinen}
\affiliation{%
 \institution{IBM Research}
 \city{Yorktown Heights} 
 \state{NY}
 \country{USA}
 \postcode{10598}}
\email{jenna.reinen@ibm.com}

\author{Irina Rish}
\affiliation{%
 \institution{Mila, Université de Montréal}
 \city{Montreal} 
 \state{Quebec}
 \country{Canada}
 \postcode{H3T}}
\email{irina.rish@mila.quebec}


\begin{abstract}

Drawing an inspiration from behavioral studies of human decision making, we propose here a more general and flexible parametric framework for  reinforcement learning that  extends  standard Q-learning to a two-stream model for processing  positive and negative rewards, and  allows to  incorporate a wide range of reward-processing biases --  an  important component of human decision making which can help us better understand a wide spectrum of multi-agent interactions in complex real-world socioeconomic systems, as well as various neuropsychiatric conditions associated with disruptions in normal reward processing. From the computational perspective, we observe that the proposed {\em Split-QL} model and its clinically inspired variants consistently outperform standard Q-Learning and SARSA methods, as well as recently proposed Double Q-Learning approaches, on simulated tasks with particular reward distributions, a real-world dataset capturing human decision-making in gambling tasks, and the Pac-Man game in a lifelong learning setting across different reward stationarities. 
\footnote{The codes to reproduce all the experimental results can be accessed at \url{https://github.com/doerlbh/mentalRL}. Full text can be accessed at \url{https://arxiv.org/abs/1906.11286}.}
\end{abstract}


\maketitle
\vspace{-0.2cm}


\section{Introduction}

In order to better model and understand human decision-making behavior, scientists usually investigate reward processing mechanisms in healthy subjects \cite{perry2015reward}. However, neurodegenerative and psychiatric disorders, often associated with reward processing disruptions, can provide an additional resource for deeper understanding of human decision making mechanisms. Furthermore, from the perspective of evolutionary psychiatry, various mental disorders, including depression, anxiety, ADHD, addiction and even schizophrenia can be considered as ``extreme points'' in a continuous spectrum of behaviors and traits developed for various purposes during evolution, and  somewhat less extreme versions of those traits can be actually beneficial in specific environments (e.g., ADHD-like fast-switching attention can be life-saving in certain environments, etc.). 
Thus, modeling decision-making biases and disorder-relevant traits may enrich the existing computational decision-making models, leading to potentially more flexible and better-performing algorithms.

In this paper, we  build upon the standard Q-Learning (QL), a state-of-art RL approach, and extend it to a parametric family of models, called {\em Split-QL},  where the reward information is split into two streams,  positive and negative. The model puts different weight parameters on the incoming positive and negative rewards, and  imposes different discounting factors on  positive and negative reward accumulated in the past. This simple but powerful extension of QL allows to capture a variety of  reward-processing biases observed in human behavior. In particular, our model was loosely inspired by a several well-known reward processing imbalances associated with certain neuropsychiatric conditions \footnote{For example, it was shown that (unmedicated) patients with  Parkinson's disease  learn better from negative  rewards rather than from positive ones \cite{frank2004carrot}, and that  addictive behaviors  may be associated with an inability to forget strong stimulus-response associations from the past (unable to properly discount the past rewards) \cite{redish2007reconciling}.
} and our attempt to capture at least certain aspects of such imbalances within a single computational model. Our empirical evaluation involved a range of Split-QL models with a variety of parameter combinations reflecting different biases, as well as baseline approaches including QL and SARSA \cite{rummery1994line}, as well as a closely related to our work Double Q-learning (DQL) \cite{hasselt2010double}. We show that the split models competitively outperform the QL, SARSA, as well as DQL, on the artificial data simulated from a known Markov Decision Process (MDP), and on the Iowa Gambling Task (IGT) \cite{steingroever2015data} -- a real-life dataset reflecting human decision-making on a gambling task. In the experiment of the PacMan game in a lifelong learning setting where the reward distributions changes in a stochastic process over multiple stages of learning, the Split-QL models demonstrated a clear advantage over baselines with respect to the performance and adaptation to the new environments. While further and more extensive empirical evaluation may be required in order to identify the conditions when the proposed approach is likely to outperform state-of-art, we hope that our preliminary empirical results already indicate the promise of the proposed approach -- a simple yet more powerful and adaptive extension of QL based on inspirations from neuropsychology and in-depth studies of human reward processing biases.

\section{Background and Related Work}
\label{sec:neuro}

\textbf{Cellular computation of reward and reward violation.} Decades of evidence has linked dopamine function to reinforcement learning via neurons in the midbrain and its connections in the basal ganglia, limbic regions, and cortex. Firing rates of dopamine neurons computationally represent reward magnitude, expectancy, and violations (prediction error) and other value-based signals \cite{Schultz1997}. This allows an animal to update and maintain value expectations associated with particular states and actions. When functioning properly, this helps an animal develop a policy to maximize outcomes by approaching/choosing cues with higher expected value and avoiding cues associated with loss or punishment. The mechanism is conceptually similar to  reinforcement learning widely used in computing and robotics \cite{Sutton1998}, suggesting mechanistic overlap in humans and AI. Evidence of Q-learning and actor-critic models have been observed in spiking activity in midbrain dopamine neurons in primates \cite{Bayer2005} and in the human striatum using the BOLD signal \cite{ODoherty2004}. 

\textbf{Positive vs. negative learning signals.} Phasic dopamine signaling represents bidirectional (positive and negative) coding for prediction error signals \cite{Hart2014}, but underlying mechanisms show differentiation for reward relative to punishment learning \cite{Seymour2007}. Though representation of cellular-level aversive error signaling has been debated \cite{Dayan2008}, it is widely thought that rewarding, salient information is represented by phasic dopamine signals, whereas reward omission or punishment signals are represented by dips or pauses in baseline dopamine firing \cite{Schultz1997}. These mechanisms have downstream effects on motivation, approach behavior, and action selection. Reward signaling in a direct pathway links striatum to cortex via dopamine neurons that disinhibit the thalamus via the internal segment of the globus pallidus and facilitate action and approach behavior. Alternatively, aversive signals may have an opposite effect in the indirect pathway mediated by D2 neurons inhibiting thalamic function and ultimately action, as well \cite{Frank2006}. Manipulating these circuits through pharmacological measures has demonstrated computationally-predictable effects that bias learning from positive or negative prediction error in humans \cite{frank2004carrot}, and contribute to our understanding of perceptible differences in human decision making when differentially motivated by loss or gain \cite{Tversky1981}.
	
\textbf{Clinical Implications.} Highlighting the importance of using computational models to understand predict disease outcomes, many symptoms of neurological and psychiatric disease are related to biases in learning from positive and negative feedback \cite{Maia2011}. Human studies have shown that over-expressed reward signaling in the direct pathway may enhance the value associated with a state and incur pathological reward-seeking behavior, like gambling or substance use. Conversely, enhanced aversive error signals results in dampening of reward experience and increased motor inhibition, causing symptoms that decrease motivation, such as apathy, social withdrawal, fatigue, and depression. Further, exposure to a particular distribution of experiences during critical periods of development can biologically predispose an individual to learn from positive or negative outcomes, making them more or less susceptible to risk for brain-based illnesses \cite{Holmes2018}. These points distinctly highlight the need for a greater understanding of how intelligent systems differentially learn from rewards or punishments, and how experience sampling may impact RL during influential training. 

\label{sec:related}

\textbf{Abnormal Processing in Psychiatric and Neurological Disorders.}
The literature on the reward processing abnormalities in particular neurological and psychiatric disorders is quite extensive; below we summarize some of the recent developments in this fast-growing field.
It is well-known that the neuromodulator dopamine plays a key role in reinforcement learning processes. Parkinson's disease (PD) patients, who have depleted dopamine in the basal ganglia, tend to have impaired performance on tasks that require learning from trial and error. For example, \cite{frank2004carrot} demonstrate that off-medication PD patients are better at learning to avoid choices that lead to negative outcomes than they are at learning from positive outcomes, while dopamine medication typically used to treat PD symptoms reverses this bias. Other psychiatric and neurological conditions may present with reinforcement deficits related to non-dopaminergic mechanisms. For instance, Alzheimer's disease (AD) is the most common cause of dementia in the elderly and, besides memory impairment, it is associated with a variable degree of executive function impairment and visuospatial impairment. As discussed in \cite{perry2015reward}, AD patients have decreased pursuit of rewarding behaviors, including loss of appetite; these changes are often secondary to apathy, associated with diminished reward system activity. Furthermore, poor performance on certain tasks is correlated with memory impairments. Frontotemporal dementia (bvFTD) typically involves a progressive change in personality and behavior including disinhibition, apathy, eating changes, repetitive or compulsive behaviors, and loss of empathy \cite{perry2015reward}, and it is hypothesized that those changes are associated with abnormalities in reward processing. For example, changes in eating habits with a preference for sweet, carbohydrate rich foods and overeating in bvFTD patients can be associated with abnormally increased reward representation for food, or impairment in the negative (punishment) signal
associated with fullness.  Authors in \cite{luman2009does} suggest that the strength of the association between a stimulus and the corresponding response is more susceptible to degradation in  Attention-deficit/hyperactivity disorder (ADHD) patients, which suggests problems with storing the stimulus-response associations. Among other functions, storing the associations requires working memory capacity, which is often impaired in ADHD patients. In \cite{redish2007reconciling}, it is demonstrated that patients suffering from addictive behavior have heightened stimulus-response associations, resulting in enhanced reward-seeking behavior for the stimulus which generated such association. 
Decreased reward response may underlie a key system mediating the anhedonia and depression, which are common in chronic pain \cite{taylor2016mesolimbic}. A variety of computational models was proposed for studying the disorders of reward processing in specific disorders, including, among others \cite{frank2004carrot,seeley2012frontotemporal,hauser2016computational,dezfouli2009neurocomputational,redish2007reconciling,hess2014beyond}.
However, none of the above studies is proposing a unifying model that can represent a wide range of reward processing disorders.

\textbf{Computational Models of Reward Processing in Mental Disorders.}
A wide range of models was proposed for studying the disorders of reward processing. \cite{frank2004carrot} presented some evidence for a mechanistic account of how the human brain implicitly learns to make choices leading to good outcomes, while avoiding those leading to bad ones. Consistent results across two tasks (a probabilistic one and a deterministic one), in both medicated and non-medicated Parkinson's patients, provide substantial support for a dynamic dopamine model of cognitive reinforcement learning. In \cite{seeley2012frontotemporal}, the authors review the evolving bvFTD literature and propose a simple, testable network-based working model for understanding bvFTD. Using a computational multilevel approach, a study presented in \cite{hauser2016computational} suggests that ADHD is associated with impaired gain modulation in systems that generate increased behavioral variability. This computational, multilevel approach to ADHD provides a framework for bridging gaps between descriptions of neuronal activity and behavior, and provides testable predictions about impaired mechanisms.
Based on the dopamine hypotheses of cocaine addiction and the assumption of decreased brain reward system sensitivity after long-term drug exposure, the work by \cite{dezfouli2009neurocomputational} proposes a computational model for cocaine addiction. By utilizing average reward temporal difference reinforcement learning, this work incorporates the elevation of basal reward threshold after long-term drug exposure into the model of drug addiction proposed by \cite{redish2007reconciling}. The proposed model is consistent with the animal models of drug seeking under punishment. In the case of non-drug reward, the model explains increased impulsivity after long-term drug exposure.

In the study by \cite{hess2014beyond}, a simple heuristic model is developed to simulate individuals’ choice behavior by varying the level of decision randomness and the importance given to gains and losses. The findings revealed that risky decision-making seems to be markedly disrupted in patients with chronic pain, probably due to the high cost that pain and negative mood impose on executive control functions. Patients’ behavioral performance in decision-making tasks, such as the Iowa Gambling Task (IGT), is characterized by selecting cards more frequently from disadvantageous than from advantageous decks, and by switching more often between competing responses, as compared with healthy controls.

Under a similar high-level motivation and inspiration to our approach, MaxPain (MP) is another related algorithm introduced in \cite{elfwing2017parallel} proposing a parallel reward processing mechanism upon Q-Learning, stemming from neuroscientific studies suggesting separate reward and punishment processing systems. However, apart from this common motivation, there is a crucial difference between \cite{elfwing2017parallel}'s approach and ours: our temporal difference (TD) errors are computed entirely separately for the two streams and their corresponding $Q^+$ and $Q^-$ values, modeling two independent autonomous policies (i.e. entirely ``split''), while \cite{elfwing2017parallel}'s approach still uses the common Q value in the corresponding argmax and argmin operations for the rewards and punishments. Furthermore, we introduce discount parameters (weights) on both immediate and historic rewards and punishments, a feature not present in \cite{elfwing2017parallel}. In the empirical evaluation, we introduced MP as a baseline and confirmed this algorithmic difference by observing that its behavior differs significantly from our SQL in both tasks and performed poorly in the reinforcement learning game playing problem.

\textbf{Overall Perspective.}
To the best of our knowledge, this work is the first one to propose a generalized version of Reinforcement Learning algorithm which incorporates a range of reward processing biases associated with various mental disorders and shows how different parameter settings of the proposed model lead to behavior mimicking a wide range of impairments in multiple neurological and psychiatric disorders. Most importantly, our reinforcement learning algorithm based on generalization of Q-Learning outperforms the baseline method on multiple artificial scenarios.

To put our results in perspective, we mention that this paper is the third in a series of agents modeling mental disorders. The entry point is \cite{bouneffouf2017bandit}, where we introduced the split mechanism to the multi-armed bandit (MAB) problem \cite{Survey} where the state and context are unavailable. We demonstrated in that work, that certain types of reward bias, despite their inspirations from mental ``disorders'', can be beneficial in online learning tasks. In \cite{lin2020unified}, we further extended the two-stream reward processing to the contextual bandit (CB) problem where a context is available to the agents, and demonstrated how agents with different clinically-inspired reward bias learn from contexts in distinct ways. In this work, we extended these behavioral agents into the full reinforcement learning problem, in which we assume that the agents need to learn a policy given both the state and the context, with the algorithm initially proposed in \cite{lin2019split}. In summary, we aim to unify all three levels as a parametric family of models, where the reward information is split into two streams,  positive and negative.

\section{Problem Setting}
\label{sec:problem}
\subsection{Reinforcement Learning}	\label{subsec:rl}
Reinforcement learning defines a class of algorithms for solving problems modeled as Markov decision processes (MDP) \cite{Sutton1998}. An MDP is defined by the tuple $(\mathcal{S}, \mathcal{A}, \mathcal{T}, \mathcal{R}, \gamma)$, where  $\mathcal{S}$ is a set of possible states, $\mathcal{A}$ is a set of actions, $\mathcal{T}$ is a transition function defined as $\mathcal{T}(s, a, s')=\Pr(s'\vert s,a)$, where $s, s'\in \mathcal{S}$ and $a\in \mathcal{A}$, and $\mathcal{R}: \mathcal{S}\times \mathcal{A} \times \mathcal{S}\mapsto \mathbb{R}$ is a reward function, $\gamma$ is a discount factor that decreases the impact of the past reward on current action choice.
 
Typically,  the objective is to maximize the discounted long-term reward, assuming  an infinite-horizon decision process, i.e. to find 
a policy function $\pi: \mathcal{S} \mapsto \mathcal{A}$ which specifies the action to take in a  given state, so that the cumulative reward is maximized:
\vspace{-0.1in}
\begin{align}
\max_{\pi} \sum_{t=0}^{\infty}\gamma^t \mathcal{R}(s_t,a_t, s_{t+1}).
\end{align}
 One way to solve this problem is by using the  {\em Q-learning approach} with function approximation \cite{bertsekas1996neuro}. The Q-value of a state-action pair, $\mathcal{Q}(s,a)$, is the expected future discounted reward for taking action $a \in \mathcal{A}$ in state $s \in \mathcal{S}$. A common method to handle very large state spaces is to approximate the $\mathcal{Q}$ function as a linear function of some features. Let $\boldsymbol{\psi}(s,a)$ denote relevant features of the state-action pair $\langle s, a \rangle$. Then, we assume $\mathcal{Q}(s,a) = \boldsymbol{\theta} \cdot \boldsymbol{\psi}(s,a)$, where $\boldsymbol{\theta}$ is an unknown vector to be learned by interacting with the environment. Every time the reinforcement learning agent takes action $a$ from state $s$, obtains immediate reward $r$ and reaches new state $s'$, the parameter $\boldsymbol{\theta}$ is updated using
\begin{equation} \label{eqn:q-update}
\begin{aligned}
\text{difference} &= \left[r + \gamma \max_{a'} \mathcal{Q}(s',a')\right] - \mathcal{Q}(s,a)\\
\theta_i &\gets \theta_i + \alpha \cdot \text{difference} \cdot \psi_i(s,a),
\end{aligned}
\end{equation}
where $\alpha$ is the learning rate. A common approach to the exploration vs. exploitation trade-off is to use the  $\epsilon$-greedy strategy; namely, during the training phase, a random action is chosen with the probability   $\epsilon$,  while the optimal action maximizing  Q-value is chosen with the probability $1-\epsilon$. The agent executes this strategy, updating  its $\boldsymbol{\theta}$ parameter according to Equation~\eqref{eqn:q-update},  until  the Q-value converges, or some maximal number of time-steps is reached.

\section{Split Q-Learning: Two-Stream Approach to RL}
\label{sec:method}

We now extend  Q-learning to a more flexible  framework, inspired by a wide range of reward-processing biases discussed above. The proposed {\em Split Q-Learning } (Algorithm \ref{alg:SQL}) treats  positive and negative rewards in two separate streams. It introduces four hyper-parameters which represent, for both positive and negative  streams, the reward processing weights (biases), as well as discount factors for the past rewards:  $\lambda_+ $ and $\lambda_-$ are the discount factors applied to the previously accumulated positive and negative rewards, respectively, while $w_+$ and $w_-$ represent the weights on the positive and negative rewards at the current iteration.      $Q^{+}$,  $Q^{-}$ and $Q$ are the Q-value   tables   recording   positive, negative and total feedbacks, respectively. We assume that at each step, an agent receives both positive and negative rewards, denote $r^+$ and $r^-$, respectively (either one of them can be zero, of course).\footnote{A different formulation under the same motivation is explored and included in as the algorithm \ref{alg:SQL2} in Appendix \ref{sec:theOtherSQL}, with a similar numerical results.}

\begin{algorithm}[th]
 \caption{Split Q-Learning}
\label{alg:SQL}
\begin{algorithmic}[1]
  \STATE {\bfseries } {\bf Initialize:} $Q$, $Q^+$, $Q^-$ tables (e.g., to all zeros)
  \STATE \textbf{For} each episode $e$ \textbf{do}
 \STATE {\bfseries } \quad Initialize state $s$
 \STATE {\bfseries } \quad \textbf{Repeat} for each step $t$ of the episode $e$
 \STATE {\bfseries }  \quad \quad  $Q(s,a') := Q^{+}(s,a') +  Q^{-}(s,a'), \forall a' \in A_t$
 \STATE {\bfseries } \quad \quad Take action $a = \arg \max_{a'}Q(s,a')$, and
  \STATE {\bfseries } \quad \quad Observe $s'\in S$, $r^+ \text{ and } r^- \in R(s)$ \\
  \STATE {\bfseries } \quad \quad $s \leftarrow s'$ \\
 \STATE {\bfseries }  \quad \quad $Q^{+}(s,a):=\lambda_+\hat{Q}^{+}(s,a)+ $ \\
  \quad \quad
 $\alpha_t(w_{+}r^{+}+\gamma \max_{a'}\hat{Q}^{+}(s',a')-\hat{Q}^{+}(s,a))$
 \STATE {\bfseries } \quad \quad $Q^{-}(s,a):= \lambda_-\hat{Q}^{-}(s,a)+$ \\
 \quad \quad
$ \alpha_t(w_{-}r^{-}+\gamma   \max_{a'}\hat{Q}^{-}(s',a')-\hat{Q}^{-}(s,a))$
 \STATE {\bfseries } \quad \textbf{until} s is the terminal state
 \STATE {\bfseries }\textbf{End for}
 \end{algorithmic}
\end{algorithm}

\subsection{Biases in Reward Processing Models }
\label{subsec:mental}
In this section we describe how specific constraints on the model parameters in the proposed algorithm can generate a range of reward processing biases, and introduce several  instances of the split-QL model associated with those biases; the corresponding parameter settings are presented in  Table \ref{tab:parameter}. As we demonstrate later, specific biases may be actually beneficial in some settings, and our parameteric approach often outperforms the standard Q-learning due to  increased generality and flexibility of our two-stream, multi-parametric formulation.


Note that the {\em standard} split-QL (SQL) approach correspond to setting the four (hyper)parameters used in our model to 1. We also introduce two variants which only learn from one  of the two reward streams: negative Q-Learning (NQL)  and positive Q-Learning (PQL), by setting to zero   $ \lambda_+$ and $w_+$, or  $ \lambda_-$ and $w_-$, respectively. Next, we introduce the model which incorporates some mild forgetting of the past rewards or losses (0.5 weights) and calibrating the other models with respect to this one; we refer to this model as M for ``moderate'' forgetting. 


\begin{table}[tb]
\centering
\caption{Parameter settings for different types of reward biases in the Split-QL model.}
\label{tab:parameter}
\resizebox{0.95\columnwidth}{!}{
 \begin{tabular}{ l | c | c | c | c }
  & $\lambda_+$   & $w_+$     & $\lambda_-$      & $w_-$ \\ \hline
  ``Addiction'' (ADD)   & $1 \pm 0.1$   & $1 \pm 0.1$    & $0.5 \pm 0.1$   & $1 \pm 0.1$ \\
  ``ADHD''  & $0.2\pm 0.1$  & $1 \pm 0.1$    & $0.2 \pm 0.1$   & $1 \pm 0.1$ \\
  ``Alzheimer's'' (AD)   & $0.1 \pm 0.1$  & $1 \pm 0.1$    & $0.1 \pm 0.1$   & $1 \pm 0.1$ \\
  ``Chronic pain'' (CP)   & $0.5 \pm 0.1$  & $0.5 \pm 0.1$    & $1 \pm 0.1$    & $1 \pm 0.1$ \\
  ``bvFTD''  & $0.5 \pm 0.1$  & $100 \pm 10$    & $0.5 \pm 0.1$   & $1 \pm 0.1$ \\
  ``Parkinson's'' (PD)   & $0.5 \pm 0.1$  & $1 \pm 0.1$    & $0.5\pm 0.1$   & $100\pm 10$\\
  ``moderate'' (M)   & $0.5 \pm 0.1$  & $1 \pm 0.1$    & $0.5 \pm 0.1$   & $1 \pm 0.1$ \\
  \hline
  Standard Split-QL (SQL)  & 1        & 1         & 1         & 1\\
  Positive Split-QL (PQL) & 1        & 1         & 0         & 0\\
  Negative Split-QL (NQL)  & 0        & 0         & 1         & 1\\
 \end{tabular}
 }
    \vspace{0.1in}
 \par\caption*{For each agent, we set the four parameters: $\lambda_+$ and $\lambda_-$ as the weights of the previously accumulated positive and negative rewards, respectively, $w_+$ and $w_-$ as the weights on the positive and negative rewards at the current iteration.}
     \vspace{-0.2in}
 \end{table}

We will now introduced several models inspired by certain reward-processing biases in a range of mental disorders-like behaviors in table \ref{tab:parameter} \footnote{DISCLAIMER: while we use disorder names for the models, we are not claiming that the models accurately capture all aspects of the  corresponding disorders.}. Recall that PD patients are typically better at learning to avoid negative outcomes than at learning to achieve positive outcomes \cite{frank2004carrot}; one way to model this is to over-emphasize negative rewards, by placing a high weight on them, as compared to the reward processing in healthy individuals. Specifically, we will assume the parameter $w_-$ for PD patients to be much higher than normal $w_-$ (e.g., we use $w_-=100$ here), while the rest of the parameters will be in the same range for both healthy and PD individuals.
Patients with bvFTD are prone to overeating which may represent increased reward representation. To model this impairment in bvFTD patients, the parameter of the model could be modified as follow: $w_+^M <<w_+$ (e.g.,  $w_+=100$ as shown in Table \ref{tab:parameter}), where $w_+$ is the parameter of the bvFTD model has, and the rest of these parameters are equal to the normal one.
To model apathy in patients with Alzheimer's, including downplaying rewards and losses, we will assume that the parameters $\lambda_+$ and $\lambda_-$ are  somewhat smaller than normal, $ \lambda_+ < \lambda_+^M$ and $\lambda_- < \lambda_-^M $ (e.g,  set to 0.1 in Table \ref{tab:parameter}), which models the tendency to  forget both positive and negative rewards.
Recall that ADHD may be involve impairments in storing stimulus-response associations. In our ADHD model, the parameters $\lambda_+$ and $\lambda_-$ are smaller than normal, $\lambda_+^M > \lambda_+$ and $\lambda_-^M > \lambda_-$, which models forgetting of both positive and negative rewards. Note that while this model appears similar to Alzheimer's model described above, the forgetting factor will be  less pronounced, i.e. the $\lambda_+$ and $\lambda_-$ parameters are larger than those of the  Alzheimer's model (e.g., 0.2 instead of 0.1, as shown in Table  \ref{tab:parameter}).
As mentioned earlier, addiction is associated with inability to properly forget (positive) stimulus-response associations; we model this by setting the weight on previously accumulated positive reward (``memory'' )  higher than normal, $\tau >\lambda_+^M $, e.g. $\lambda_+ = 1$, while $\lambda_+^M = 0.5$. We model the reduced responsiveness to rewards in chronic pain by setting $ w_+ < w_+^M $ so there is a decrease in the reward representation, and $\lambda_- > \lambda_-^M $ so the negative rewards are not forgotten (see table \ref{tab:parameter}).

Of course, the above models should be treated only as first approximations of the reward processing biases in mental disorders, since the actual changes in reward processing are much more complicated, and the parameteric setting must be learned from actual patient data, which is a non-trivial direction for future work. Herein, we simply consider those models as specific variations of our general method, inspired by certain aspects of the corresponding diseases, and focus primarily on the computational aspects of our algorithm, demonstrating that the proposed parametric extension of QL can learn better than the baselines due to added flexibility.

\section{Empirical Results}
\label{sec:results}

\begin{table}[tb]
\begin{minipage}{\linewidth}
      \caption{\textbf{Standard agents:} MDP with 100 randomly generated scenarios of Bi-modal reward distributions}
      \label{tab:MDP} 
      \centering
      \resizebox{1\linewidth}{!}{
 \begin{tabular}{ l | c | c | c | c | c | c | c | c }
 &\multicolumn{3}{c}{Baseline} \vline & \multicolumn{5}{c}{Variants of SQL} \\
  &  QL & DQL & SARSA & SQL-alg1 & SQL-alg2 & MP & PQL & NQL \\ \hline
QL & - & \textbf{62}:38 & \textbf{55}:45 & \textbf{63}:37 & \textbf{54}:46 & 47:\textbf{53} & \textbf{65}:35 & \textbf{90}:10\\
DQL & 38:\textbf{62} & - & 40:\textbf{60} & 48:\textbf{52} & 48:\textbf{52} & 43:\textbf{57} & \textbf{55}:45 & \textbf{86}:14\\
SARSA & 45:\textbf{55} & \textbf{60}:40 & - & \textbf{63}:37 & \textbf{51}:49 & \textbf{52}:48 & \textbf{64}:36 & \textbf{88}:12\\
SQL & 37:\textbf{63} & \textbf{52}:48 & 37:\textbf{63} & - & 42:\textbf{58} & 26:\textbf{74} & \textbf{55}:45 & \textbf{72}:28\\
SQL2 & 46:\textbf{54} & \textbf{52}:48 & 49:\textbf{51} & \textbf{58}:42 & - & 39:\textbf{61} & \textbf{64}:36 & \textbf{72}:28\\
MP & \textbf{53}:47 & \textbf{57}:43 & 48:\textbf{52} & \textbf{74}:26 & \textbf{61}:39 & - & \textbf{66}:34 & \textbf{82}:18\\
PQL & 35:\textbf{65} & 45:\textbf{55} & 36:\textbf{64} & 45:\textbf{55} & 36:\textbf{64} & 34:\textbf{66} & - & \textbf{68}:32\\
NQL & 10:\textbf{90} & 14:\textbf{86} & 12:\textbf{88} & 28:\textbf{72} & 28:\textbf{72} & 18:\textbf{82} & 32:\textbf{68} & -\\
\hline
avg wins (\%)  & 55.05 & 45.20 & 53.41 & 40.53 & 47.98 & \textbf{55.68} & 37.75 & 17.93\\
avg ranking  & \textbf{3.64} & 4.30 & 3.78 & 4.81 & 4.32 & 3.66 & 5.21 & 6.28
 \end{tabular}
 }
    \vspace{0.1in}
\par\caption*{For each cell of $i$th row and $j$th column, the first number indicates the number of rounds the agent $i$ beats agent $j$, and the second number the number of rounds the agent $j$ beats agent $i$. The average wins of each agent is computed as the mean of the win rates against other agents in the pool of agents in the rows. The bold face indicates that the performance of the agent in column $j$ is the best among the agents, or better between the pair.}
     \vspace{-0.2in}
 \end{minipage}
 \end{table}

 \begin{table*}[tb]
  \centering
        \caption{\textbf{``Mental'' agents:} MDP with 100 randomly generated scenarios of Bi-modal reward distributions}
      \label{tab:MDPmental} 
 \begin{minipage}{.7\linewidth}
      \centering
      \resizebox{1\linewidth}{!}{
 \begin{tabular}{  c | c | c | c | c | c| c | c | c }
   & ADD & ADHD  & AD  & CP & bvFTD & PD & M &  avg wins (\%)   \\ \hline
QL & \textbf{91}:9 & \textbf{72}:28 & \textbf{86}:14 & \textbf{85}:15 & \textbf{94}:6 & \textbf{79}:21 & \textbf{81}:19 & 84.85\\
DQL & \textbf{86}:14 & \textbf{58}:42 & \textbf{83}:17 & \textbf{69}:31 & \textbf{87}:13 & \textbf{67}:33 & \textbf{76}:24 & 75.90\\
SARSA & \textbf{90}:10 & \textbf{73}:27 & \textbf{86}:14 & \textbf{80}:20 & \textbf{91}:9 & \textbf{75}:25 & \textbf{80}:20 & 82.97\\
SQL & \textbf{93}:7 & \textbf{67}:33 & \textbf{91}:9 & \textbf{72}:28 & \textbf{75}:25 & \textbf{76}:24 & \textbf{68}:32 & 78.21\\
\hline
avg wins (\%)  & 10.10 & 32.83 & 13.64 & 23.74 & 13.38 & 26.01 & 23.99
 \end{tabular}
 }
 \end{minipage}
    \vspace{0.1in}
 \par\caption*{As in table \ref{tab:MDP}, for each cell of $i$th row and $j$th column, the numbers $n:m$ indicates the number of rounds the two agents beats each other. The average wins of QL, DQL, SARSA, and SQL are computed as the mean of the win rates against other agents in the pool of agents in the columns.}
 \end{table*}
 
  \begin{table*}[tb]
\centering
\caption{\textbf{Schemes} of Iowa Gambling Task}
\label{tab:IGTschemes}
\resizebox{1\linewidth}{!}{
 \begin{tabular}{ l | c | l | c | c }
  Decks & win per card  & loss per card & expected value & scheme \\ \hline
  A (bad) & +100 & Frequent: -150 (p=0.1), -200 (p=0.1), -250 (p=0.1), -300 (p=0.1), -350 (p=0.1) & -25 & 1 \\
  B (bad) & +100 & Infrequent: -1250 (p=0.1) & -25 & 1 \\  
  C (good) & +50 & Frequent: -25 (p=0.1), -75 (p=0.1),-50 (p=0.3) & +25 & 1 \\  
  D (good) & +50 & Infrequent: -250 (p=0.1) & +25 & 1 \\   \hline
  A (bad) & +100 & Frequent: -150 (p=0.1), -200 (p=0.1), -250  (p=0.1), -300 (p=0.1), -350 (p=0.1) & -25 & 2 \\
  B (bad) & +100 & Infrequent: -1250 (p=0.1) & -25 & 2 \\  
  C (good) & +50 & Infrequent: -50 (p=0.5) & +25 & 2 \\  
  D (good) & +50 & Infrequent: -250 (p=0.1) & +25 & 2 \\  
 \end{tabular}
 } 
 \end{table*}
 Empirically, we evaluated the algorithms in three settings: a Markov Decision Process (MDP) gambling game, a real-life Iowa Gambling Task (IGT) \cite{steingroever2015data} and the PacMan game. There is considerable randomness in the reward, and predefined multimodality in the reward distributions of each state-action pairs, and as a result we will see that indeed Q-learning performs poorly. In all experiments, the discount factor $\gamma$ was set to be 0.95. The exploration is included with $\epsilon$-greedy algorithm with $\epsilon$ set to be 0.05. The learning rate was polynomial $\alpha_t(s, a) = 1/n_t(s, a)^{0.8}$, which was shown in previous work to be better in theory and in practice \cite{even2003learning}.
All experiments were performed and averaged for at least 100 runs, and over 100 or 500 steps of decision making actions from the initial state. To evaluate the performances of the algorithms, we need a scenario-independent measure which is not dependent on the specific selections of reward distribution parameters and pool of algorithms being considered. The final cumulative reward is subject to outliers because they are scenario-specific. The ranking is subject to selection bias due to different pools of algorithms being considered. The pairwise comparison of the algorithms, however, is independent of the selection of scenario parameters and selection of algorithms. For example, in the 100 randomly generated scenarios, algorithm X beats Y for $n$ times while Y beats X $m$ times. We may compare the robustness of each pairs of algorithms with the proportion $n:m$. 

\subsection{MDP example with bimodal rewards} 
\begin{figure*}[t]
\centering
    \includegraphics[width=\linewidth]{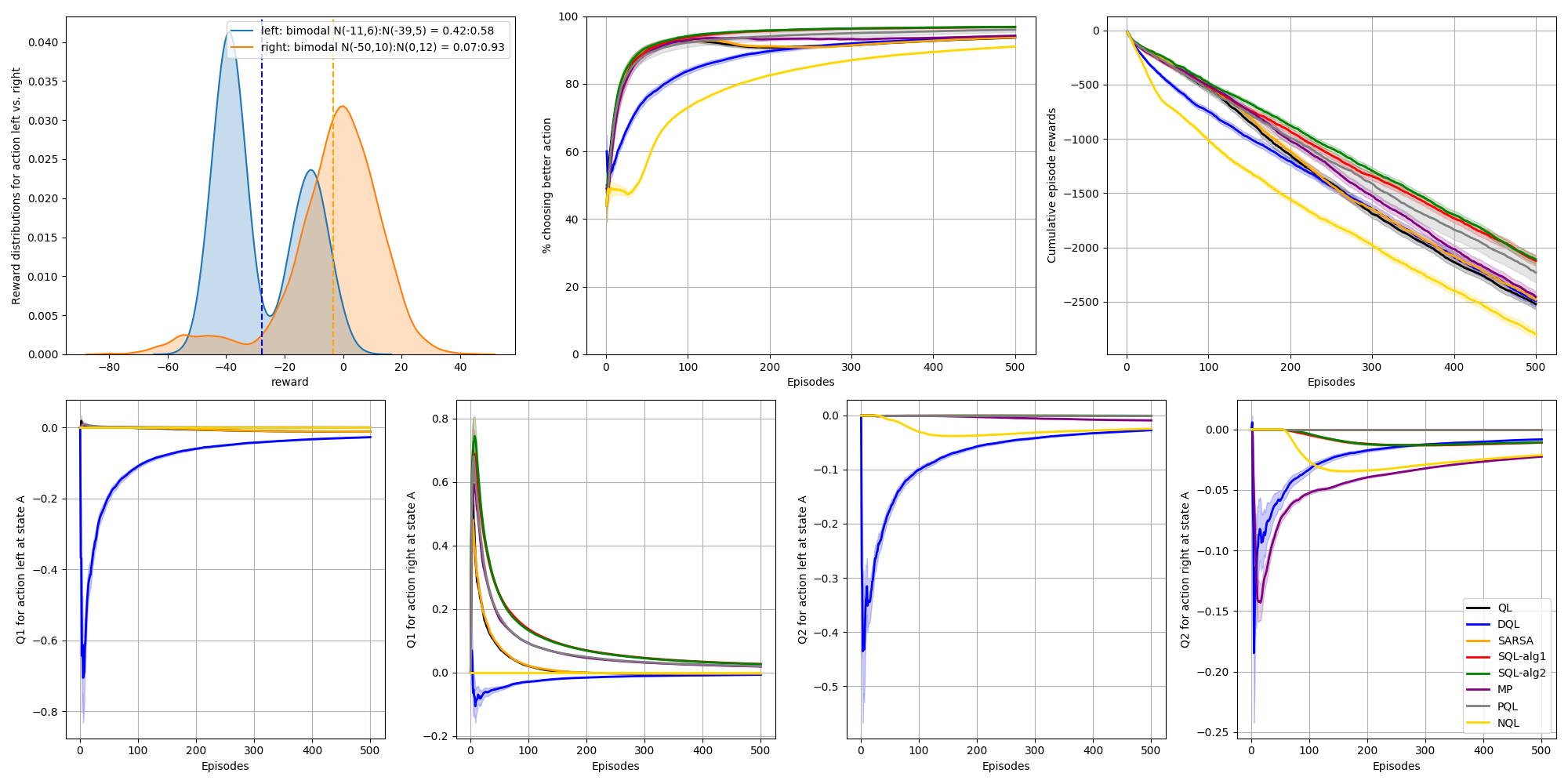}\par\caption{\textbf{Inside the two-stream learning:} example scenario where Split-QL performs better than QL and DQL.}\label{fig:MDP}
\end{figure*}

In this simple MDP example, a player starts from initial state A, choose between two actions: go left to reach state B, or go right to reach state C. Both states B and C reveals a zero rewards. From state B, the player has only one action to reach state D which reveals $n$ draws of rewards from a distribution $R_D$. From state C, the player has only one action to reach state E which reveals $n$ draws of rewards from a distribution $R_E$. The reward distributions of states D and E are both multimodal distributions (for instance, the reward $r$ can be drawn from a bi-modal distribution of two normal distributions $N(\mu=10,\sigma=5)$ with probability $p=0.3$ and $N(\mu=-5,\sigma=1)$ with $p=0.7$). In the simulations, $n$ is set to be 50. The left action (go to state B) by default is set to have an expected payout lower than the right action. However, the reward distributions can also be spread across both the positive and negative domains (as in the example shown in Figure \ref{fig:MDP}). 

Figure \ref{fig:MDP} shows an example scenario where the reward distributions, percentage of choosing the better action (go right), cumulative rewards and the changes of two Q-tables over the number of iterations, drawn with standard errors over 100 runs. Each trial consisted of a synchronous update of all 500 actions. With polynomial learning rates, we see Split Q-learning (two versions) converges the fastest and yields the highest final scores among all agents. We further observe that this performance advantage is prevalent in reward distributions with uneven bi-modes in two streams.

To better evaluate the robustness of the algorithms, we simulated 100 randomly generated scenarios of bi-modal distributions, where the reward distributions can be drawn from two normal distribution with means as random integers uniformly drawn from -100 to 100, standard deviations as random integers uniformly drawn from 0 to 20, and sampling distribution $p$ uniformly drawn from 0 to 1 (assigning $p$ to one normal distribution and $1-p$ to the other one). Each scenario was repeated 100 times.

Table \ref{tab:MDP} summarizes the pairwise comparisons between the following algorithms: Q-Learning (QL), Double Q-Learning (DQL) \cite{hasselt2010double}, State–action–reward–state–action (SARSA) \cite{rummery1994line}, MaxPain (MP) \cite{elfwing2017parallel}, Standard Split Q-Learning (SQL) with two numerical implementation (algorithm \ref{alg:SQL} and algorithm \ref{alg:SQL2} in Appendix \ref{sec:theOtherSQL}), Positive Q-Learning (PQL) and Negative Q-Learning (NQL), with the row labels as the algorithm X and column labels as algorithm Y giving $n:m$ in each cell denoting X beats Y $n$ times and Y beats X $m$ times. Among the five algorithms, RL variants with a split mechanism never seems to fail catastrophically by maintaining an overall advantages over the other algorithms (with a comparative average winning percentage). Split variants seems to benefit from the sensitivity to two streams of rewards instead of collapsing them into estimating the means as in Q-Learning.

To explore the variants of Split-QL representing different mental disorders, we also performed the same experiments on the 7 disease models proposed in section \ref{subsec:mental}. Table \ref{tab:MDP} summarizes their pairwise comparisons with SQL, SARSA, DQL and QL, where the average wins are computed averaged against three standard baseline models. Overall, ADHD, PD (``Parkinson's''), CP (``chronic pain'') and M (``moderate'') performs relatively well. 
The variation of behaviors suggest the proposed framework can potentially cover a wide spectrum of behavior by simply tuning the four hyperparameters.



\subsection{Iowa Gambling Task} 
The original Iowa Gambling Task (IGT) studies decision making where the participant needs to choose one out of four card decks (named A, B, C, and D), and can win or lose money with each card when choosing a deck to draw from \cite{bechara1994insensitivity}, over around 100 actions. In each round, the participants receives feedback about the win (the money he/she wins), the loss (the money he/she loses), and the combined gain (win minus lose). In the MDP setup, from initial state I, the player select one of the four deck to go to state A, B, C, or D, and reveals positive reward $r^+$ (the win), negative reward $r^-$ (the loss) and combined reward $r=r^++r^-$ simultaneously. Decks A and B by default is set to have an expected payout (-25) lower than the better decks, C and D (+25). For QL and DQL, the combined reward $r$ is used to update the agents. For Split-QL, PQL and NQL, the positive and negative streams are fed and learned independently given the $r^+$ and $r^-$. There are two major payoff schemes in IGT. In scheme 1, the net outcome of every 10 cards from the bad decks (decks A and B) is -250, and +250 in good decks (decks C and D). There are two decks with frequent losses (decks A and C), and two decks with infrequent losses (decks B and D). All decks have consistent wins (A and B to have +100, while C and D to have +50) and variable losses (see \ref{tab:IGTschemes}, where scheme 1 \cite{fridberg2010cognitive} has a more variable losses for deck C than scheme 2 \cite{horstmann2012iowa}).
\footnote{The raw data and descriptions of Iowa Gambling Task can be downloaded at \cite{steingroever2015data}.}

\begin{figure*}[tb]
\centering
    \includegraphics[width=\linewidth]{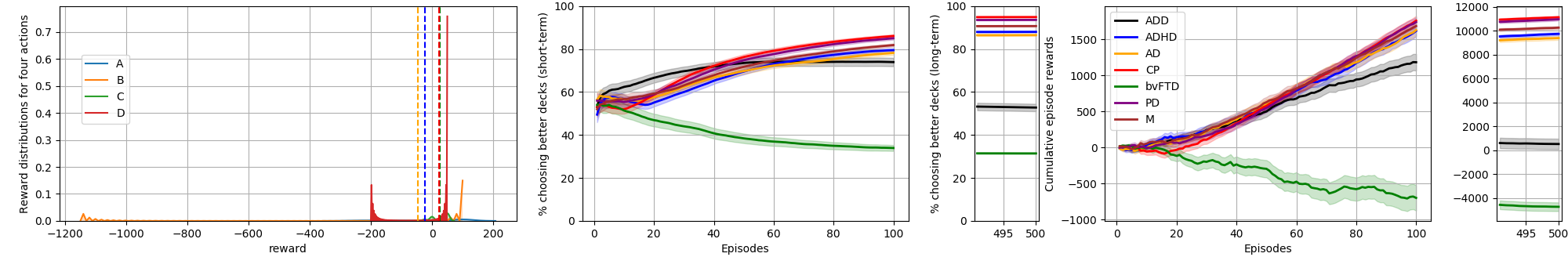}\par\caption{\textbf{Short-term vs. long-term dynamics:} learning curves of different ``mental'' agents in IGT scheme 1.}\label{fig:IGT}
\end{figure*}

\begin{figure}[tb]
\centering
    \includegraphics[width=0.43\linewidth]{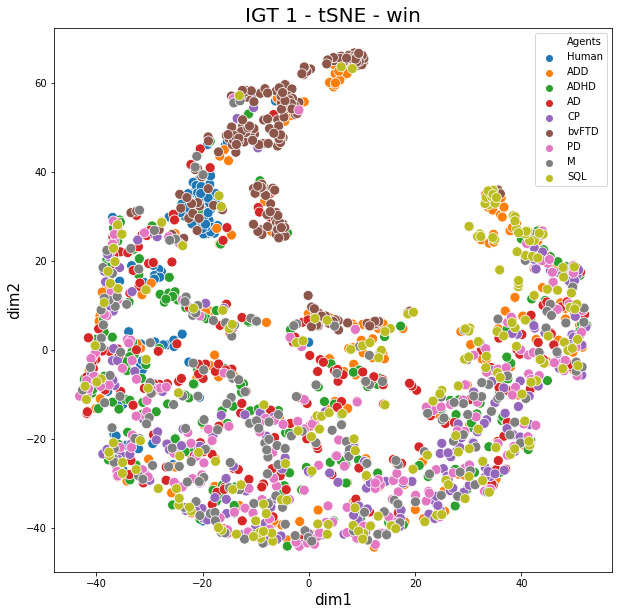}
    \includegraphics[width=0.43\linewidth]{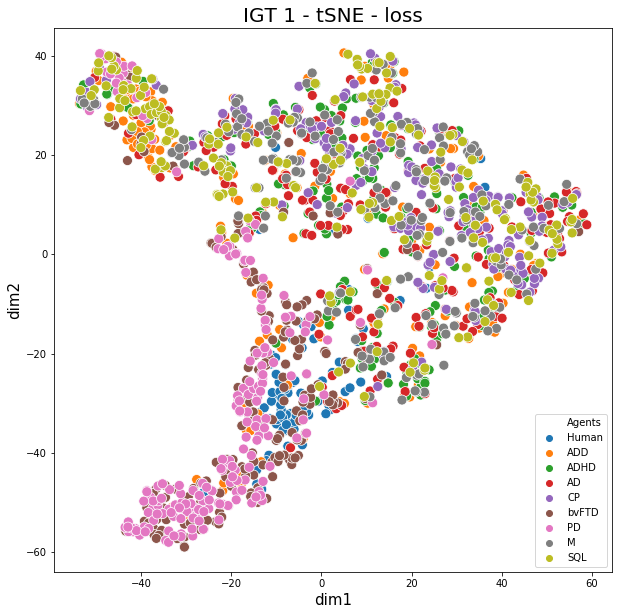}
    \par\caption{The t-SNE visualization of the behavioral trajectories of mental agents and real human data (\cite{fridberg2010cognitive,Maia2011,worthy2013decomposing}, denoted ``Human'') playing IGT scheme 1 over 95 actions: (a) the behavioral trajectories of wins, or positive reward; (b) the behavioral trajectories of losses, or negative rewards.}\label{fig:IGT_tsne}
\end{figure}

We performed the each scheme for 200 times over 500 actions. Among the variants of Split-QL and baselines QL and DQL, CP (``chronic pain'') performs best in scheme 1 with an averaged final cumulative rewards of 1145.59 over 500 draws of cards, followed by PD (``Parkinson's disease'', 1123.59). This is consistent to the clinical implication of chronic pain patients which tend to forget about positive reward information (as modeled by a smaller $\lambda_+$) and lack of drive to pursue rewards (as modeled by a smaller $w_+$). In scheme 2, PD performs best with the final score of 1129.30, followed by CP (1127.66). These examples suggest that the proposed framework has the flexibility to map out different behavior trajectories in real-life decision making (such as IGT). Figure \ref{fig:IGT} demonstrated the short-term (in 100 actions) and long-term behaviors of different mental agents, which matches clinical discoveries. For instance, ADD (``addiction'') quickly learns about the actual values of each decks (as reflected by the short-term curve) but in the long-term sticks with the decks with a larger wins (despite also with even larger losses). At around 20 actions, ADD performs significantly better than QL and DQL in learning about the decks with the better gains. Figure \ref{fig:IGT_tsne} maps the behavioral trajectories of the mental agents with real data collected from healthy human subjects playing IGT scheme 1 over 95 draws (\cite{fridberg2010cognitive,Maia2011,worthy2013decomposing}, denoted ``Human''), where we observe different clusters emerging from different reward bias. 

\subsection{Pacman game across various stationarities} 

\begin{figure}[tb]
\centering
    \includegraphics[width=0.65\linewidth]{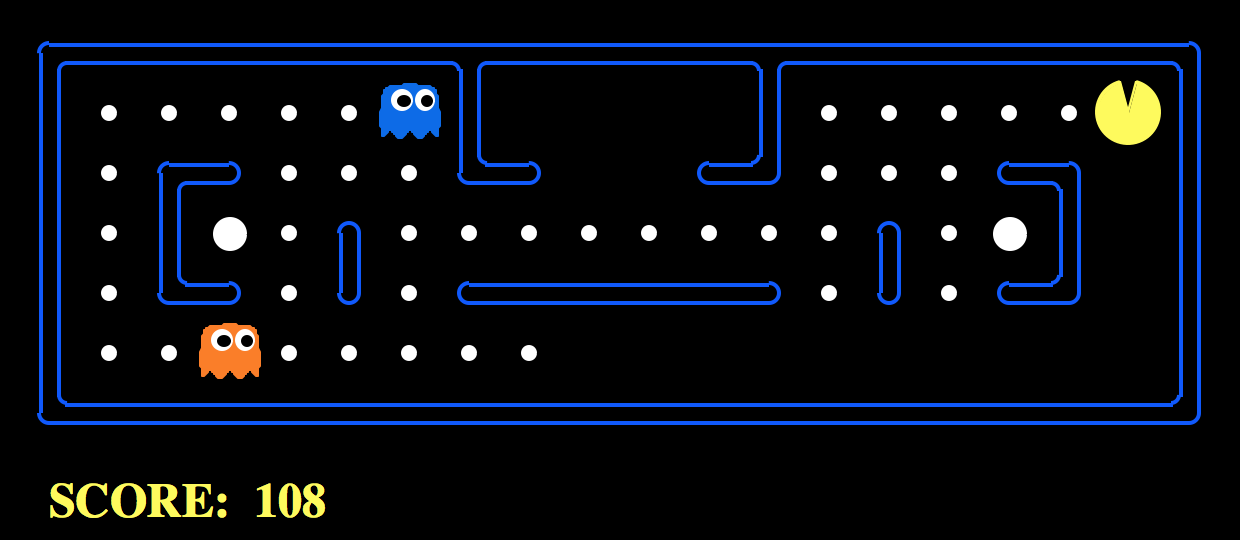}
    \par\caption{Layout of an ongoing PacMan game.}\label{fig:pacman}
\end{figure}

\begin{figure*}[tbh]
\centering
    \includegraphics[width=0.24\linewidth]{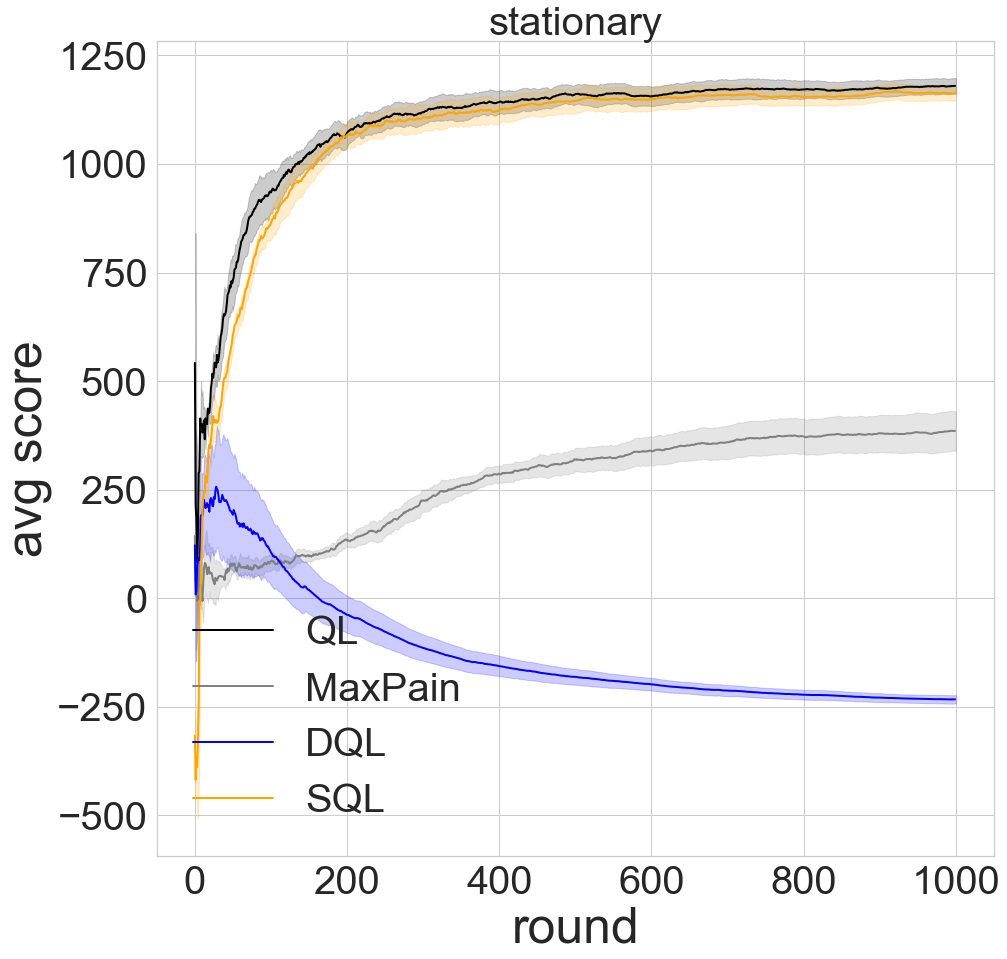}
    \includegraphics[width=0.24\linewidth]{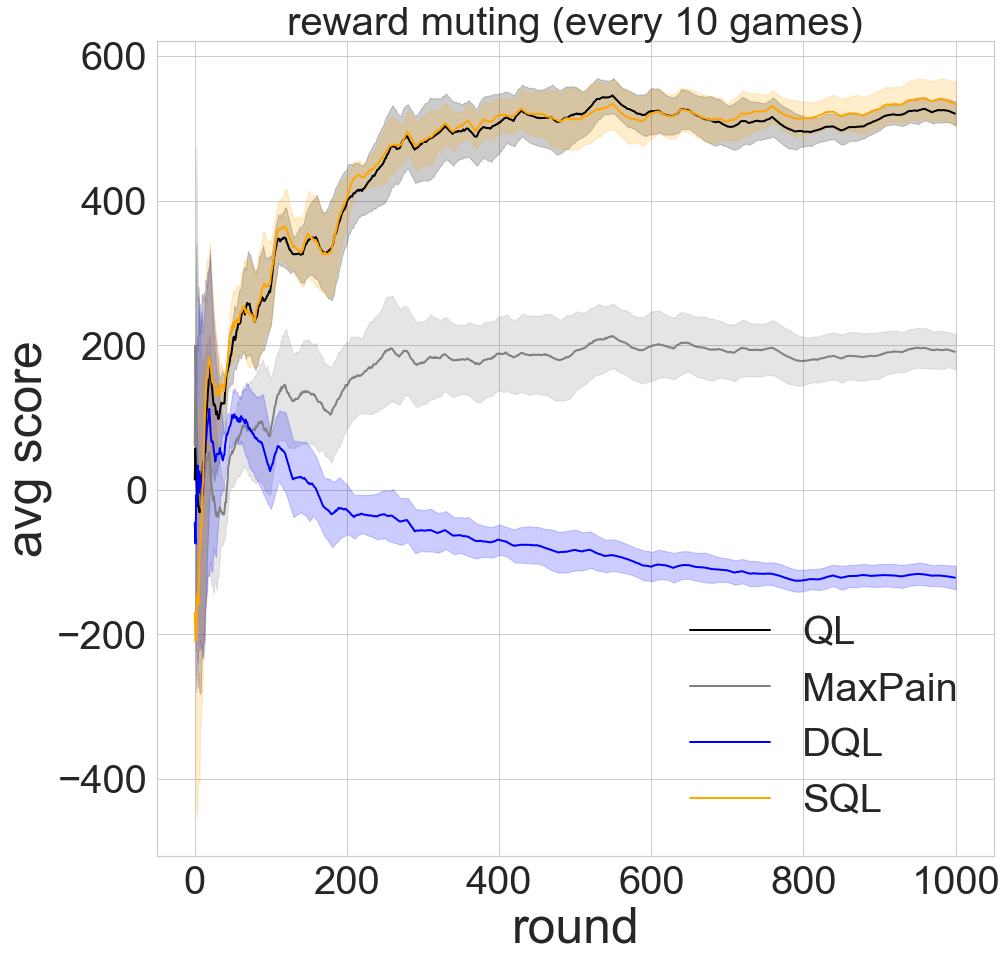}
    \includegraphics[width=0.24\linewidth]{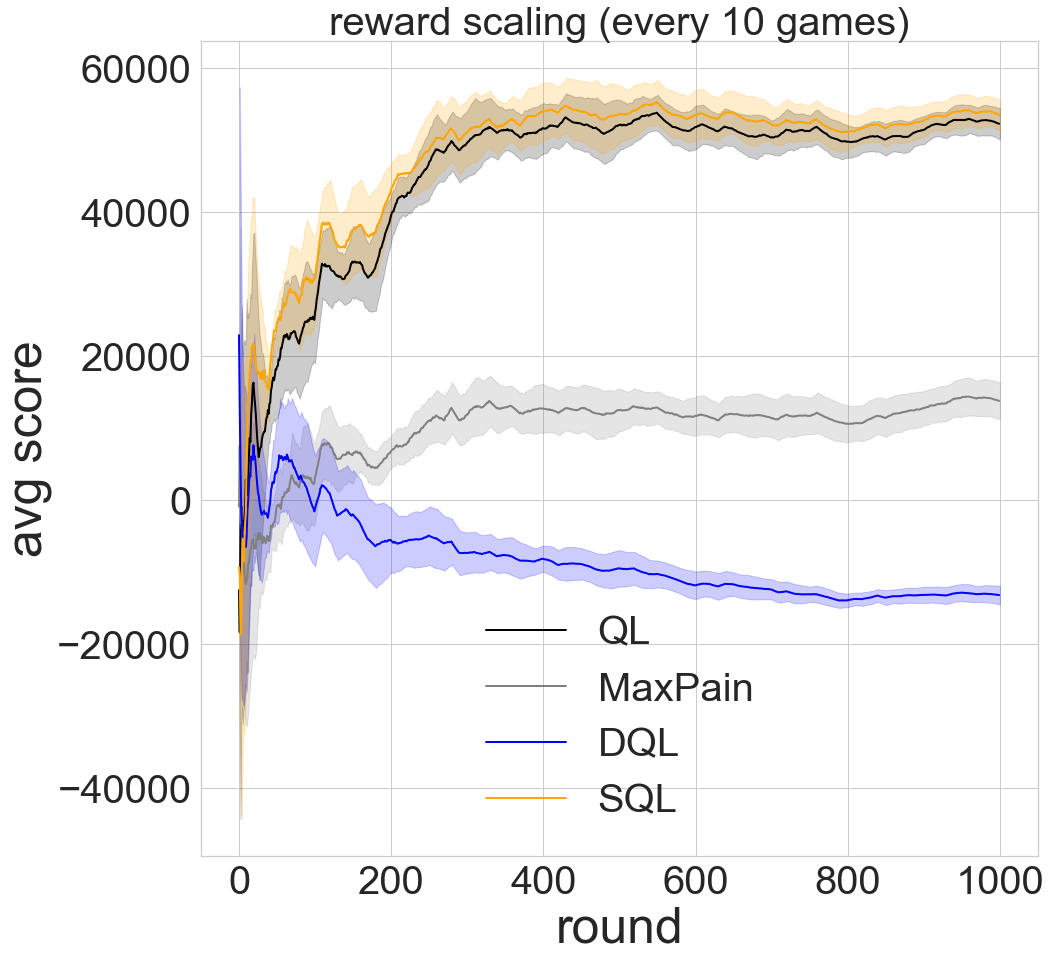}
    \includegraphics[width=0.24\linewidth]{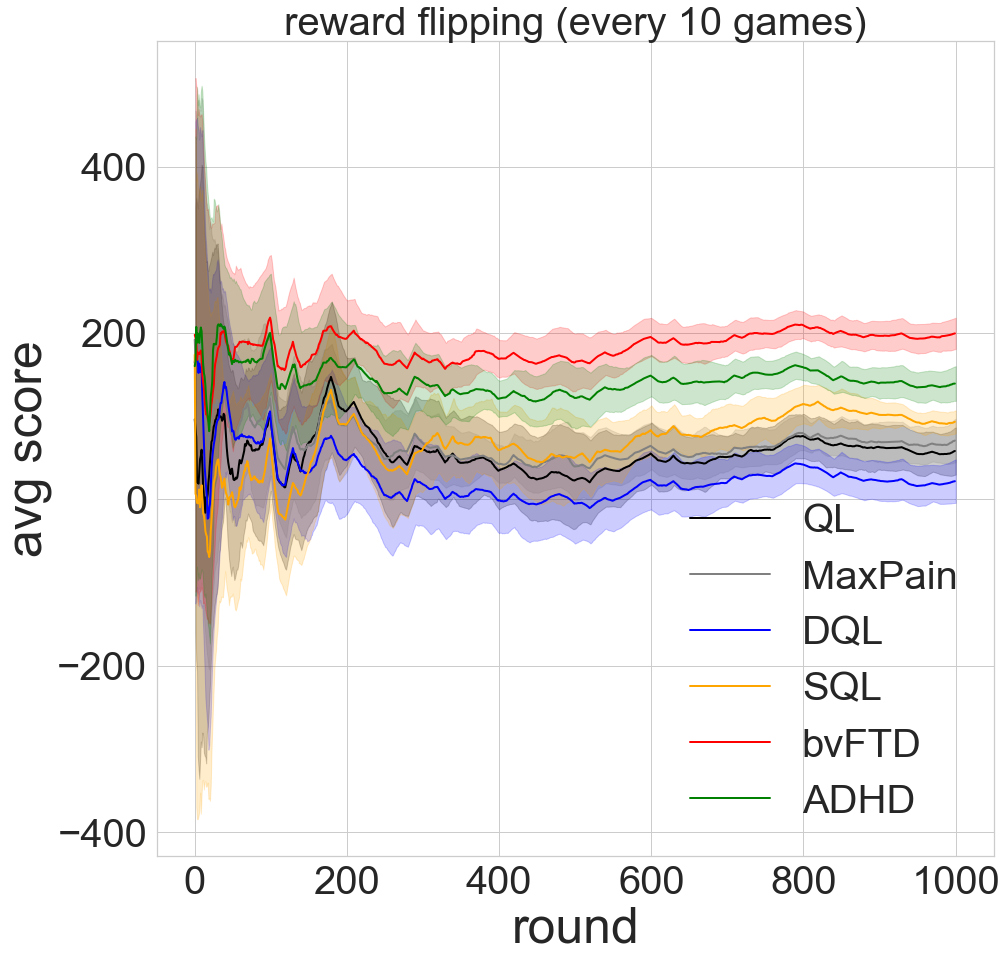}
    \par\caption{\textbf{Average final scores in Pacman game with different stationarities:} (a) stationary; (b) stochastic reward muting by every 10 rounds; (c) stochastic reward scaling by every 10 rounds; (d) stochastic reward flipping by every 10 rounds.}\label{fig:pacman_results}
\end{figure*}

We demonstrate the merits of the proposed algorithm using the classic game of PacMan (Figure \ref{fig:pacman}). The goal of the agent is to eat all the dots in the maze, known as Pac-Dots, as soon as possible while simultaneously avoiding collision with ghosts, which roam the maze trying to kill PacMan. The rules for the environment (adopted from Berkeley AI PacMan \footnote{\url{http://ai.berkeley.edu/project_overview.html}}) are as follows. There are two types of negative rewards: on collision with a ghost, PacMan loses the game and gets a negative reward of $-500$; and at each time frame, there is a constant time-penalty of $-1$ for every step taken. There are three types of positive rewards. On eating a Pac-Dot, the agent obtains a reward of $+10$. On successfully eating all the Pac-Dots, the agent wins the game and obtains a reward of $+500$. The game also has two special dots called Power Pellets in the corners of the maze, which on consumption, give PacMan the temporary ability of “eating” ghosts. During this phase, the ghosts are in a ``scared'' state for 40 frames and move at half their speed. On eating a ``scared'' ghost, the agent gets a reward of $+200$, the ghost returns to the center box and returns to its normal ``unscared'' state. As a more realistic scenarios as real-world agents, we define the agents to receive their rewards in positive and negative streams separately. Traditional agents will sum the two streams as a regular reward, while Split-QL agents will use the two streams separately.

We created several types of stationarities using the PacMan game as in \cite{lin2020diabolic}. In order to simulate a lifelong learning setting, we assume that the environmental settings arrive in batches (or stages) of episodes, and the specific rule of the games (i.e., the reward distributions) may change across those batches, while remaining stationary within each batch. The change is defined by a stochastic process of the game settings that an event $A$ is defined for the positive stream and an event $B$ is defined for the negative stream, independent of each other ($A \perp B$). In our experiment, the stochastic process is resampled every 10 rounds (i.e. a batch size of 10). 

\textbf{Stochastic reward muting.} To simulate the changes of turning on or off of a certain reward stream, we define the event $A$ as turning off the positive reward stream (i.e. all the positive rewards are set to be zero) and the event $B$ as turning off the negative reward stream (i.e. all the penalties are set to be zero). We set $\mathbb{P}(A)=\mathbb{P}(B)=0.5$.

\textbf{Stochastic reward scaling.} To simulate the changes of scaling up a certain reward stream, we define the event $A$ as scaling up the positive reward stream by 100 (i.e. all the positive rewards are multiplied by 100) and the event $B$ as scaling up the negative reward stream (i.e. penalties multiplied by 100). We set $\mathbb{P}(A)=\mathbb{P}(B)=0.5$.

\textbf{Stochastic reward flipping.} To simulate the changes of flipping certain reward stream, we define the event $A$ as flipping the positive reward stream (i.e. all the positive rewards are multiplied by -1 and considered penalties) and the event $B$ as flipping the negative reward stream (i.e. all the penalties are multiplied by -1 and considered positive rewards). We set $\mathbb{P}(A)=\mathbb{P}(B)=0.5$.

We ran the proposed agents across these different stationarities for 1000 episodes over multiple runs and plotted their average final scores with their standard errors (Figure \ref{fig:pacman_results}). In all four scenarios, Split-QL is constantly outperforming QL and DQL. It is also worth noting that in the reward flipping scenario, several mental agents are even more advantageous than the standard Split-QL as in Figure \ref{fig:pacman_results}(d), which matches clinical discoveries and the theory of evolutionary psychiatry. For instance, ADHD-like fast-switching attention seems to be especially beneficial in this very non-stationary setting of flipping reward streams. Even in a full stationary setting, the behaviors of these mental agents can have interesting clinical implications. For instance, the video of a CP (``chronic pain'') agent playing PacMan shows a clear avoidance behavior to penalties by staying at a corner very distant from the ghosts and a comparatively lack of interest to reward pursuit by not eating nearby Pac-Dots, matching the clinical characters of chronic pain patients. From the video, we observe that the agent ignored all the rewards in front of it and spent its life hiding from the ghosts, trying to elongate its life span at all costs, even if that implies a constant time penalty to a very negative final score. (The videos of the mental agents playing PacMan after training for 1000 episodes can be accessed here\footnote{\url{https://github.com/doerlbh/mentalRL/tree/master/video}}).

\section{Conclusions}
\label{sec:conclusion}


This research proposes a novel parametric family of algorithms for RL problem, extending the classical Q-Learning to model a wide range of potential reward processing biases. Our approach draws an inspiration from extensive literature on decision-making behavior in neurological and psychiatric disorders stemming from disturbances of the reward processing system, and demonstrates high flexibility of our multi-parameter model which allows to tune the weights on incoming two-stream rewards and memories about the prior reward history. Our preliminary results support multiple prior observations about reward processing biases in a range of mental disorders, thus indicating the potential of the proposed model and its future extensions to capture reward-processing aspects across various neurological and psychiatric conditions. 

The contribution of this research is two-fold: from the machine-learning perspective, we propose a simple yet powerful and more adaptive approach to RL, outperforming state-of-art QL for certain reward distributions; from the neuroscience perspective, this work is the first attempt at a general, unifying model of reward processing and its disruptions across a wide population including both healthy subjects and those with mental disorders, which has a potential to become a useful computational tool for neuroscientists and psychiatrists studying such disorders. Among the directions for future work, we plan to investigate the optimal parameters in a series of computer games evaluated on different criteria, for example, longest survival time vs. highest final score. Further work includes exploring the multi-agent interactions given different reward processing bias. These discoveries can help build more interpretable real-world RL systems. On the neuroscience side, the next steps would include further extending the proposed model to better capture observations in modern literature, as well as testing the model on both healthy subjects and patients with specific mental conditions.

\clearpage\newpage
\bibliographystyle{ACM-Reference-Format}  
\bibliography{main}  

\clearpage
\newpage


\appendix
\onecolumn

\section{\textit{SQL-2}: a different formulation}
\label{sec:theOtherSQL}

Algorithm \ref{alg:SQL2} is another version for the \textit{Split-QL} where the $w_+$ and $w_-$ are installed in the action selection step, instead of the reward perception as in algorithm \ref{alg:SQL}. The two variants are motivated by the same neurobiological inspirations, despite different algorithmic formulation.

\begin{algorithm}[tbh]
 \caption{Split Q-Learning - version 2}
\label{alg:SQL2}
\begin{algorithmic}[1]
  \STATE {\bfseries } {\bf Initialize} $Q$, $Q^+$, $Q^-$ tables (e.g., to all zeros)
  \STATE \textbf{For} each episode t \textbf{do}
 \STATE {\bfseries } \quad Initialize state $s$
 \STATE {\bfseries } \quad \textbf{Repeat} for each step of the episode $t$
 \STATE {\bfseries }  \quad \quad  $Q(s,a) := w_+ Q^{+}(s,a) + w_- Q^{-}(s,a)$
 \STATE {\bfseries } \quad \quad take action $i_t= \arg \max_{i}Q(s,i)$, and
  \STATE {\bfseries } \quad \quad observe $s'\in S$, $r^+ \text{ and } r^- \in R(s)$ \\
  \STATE {\bfseries } \quad \quad $s \leftarrow s'$ \\
 \STATE {\bfseries }  \quad \quad $Q^{+}(s,a):=\lambda_+\hat{Q}^{+}(s,a)+ $ \\
  \quad \quad
 $\alpha_t(r^{+}+\gamma \max_{a'}\hat{Q}^{+}(s',a')-\hat{Q}^{+}(s,a))$
 \STATE {\bfseries } \quad \quad $Q^{-}(s,a):= \lambda_-\hat{Q}^{-}(s,a)+$ \\
 \quad \quad
$ \alpha_t(r^{-}+\gamma   \max_{a'}\hat{Q}^{-}(s',a')-\hat{Q}^{-}(s,a))$
 \STATE {\bfseries } \quad \textbf{until} s is the terminal state
 \STATE {\bfseries }\textbf{End for}
 \end{algorithmic}
\end{algorithm}
  \vspace{-0.1in}

\section{Additional results}
\label{sec:moreResults}

\begin{figure}[h]
\centering
    \includegraphics[width=0.4\linewidth]{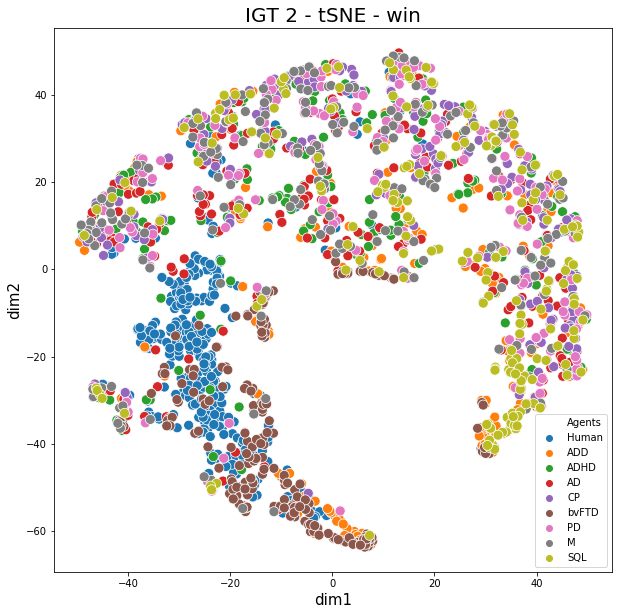}
    \includegraphics[width=0.4\linewidth]{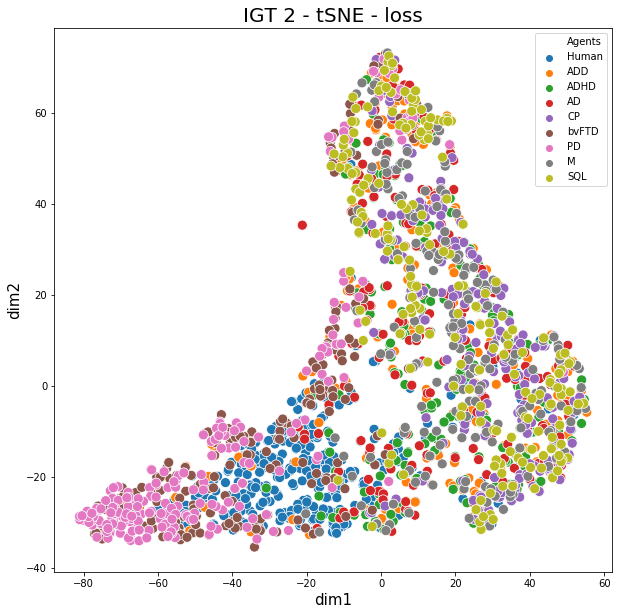}
    \par\caption{t-SNE visualization of the behavioral trajectories of mental agents and real human data (\cite{steingroever2013performance,wetzels2010bayesian}, denoted ``Human'') playing IGT scheme 2 over 95 actions: (a) trajectories of wins, or positive reward; (b) trajectories of losses, or negative rewards.}\label{fig:IGT_tsne_more}
\end{figure}

\begin{figure}[tbh]
\centering
    \includegraphics[width=0.24\linewidth]{pacman.png}
    \includegraphics[width=0.24\linewidth]{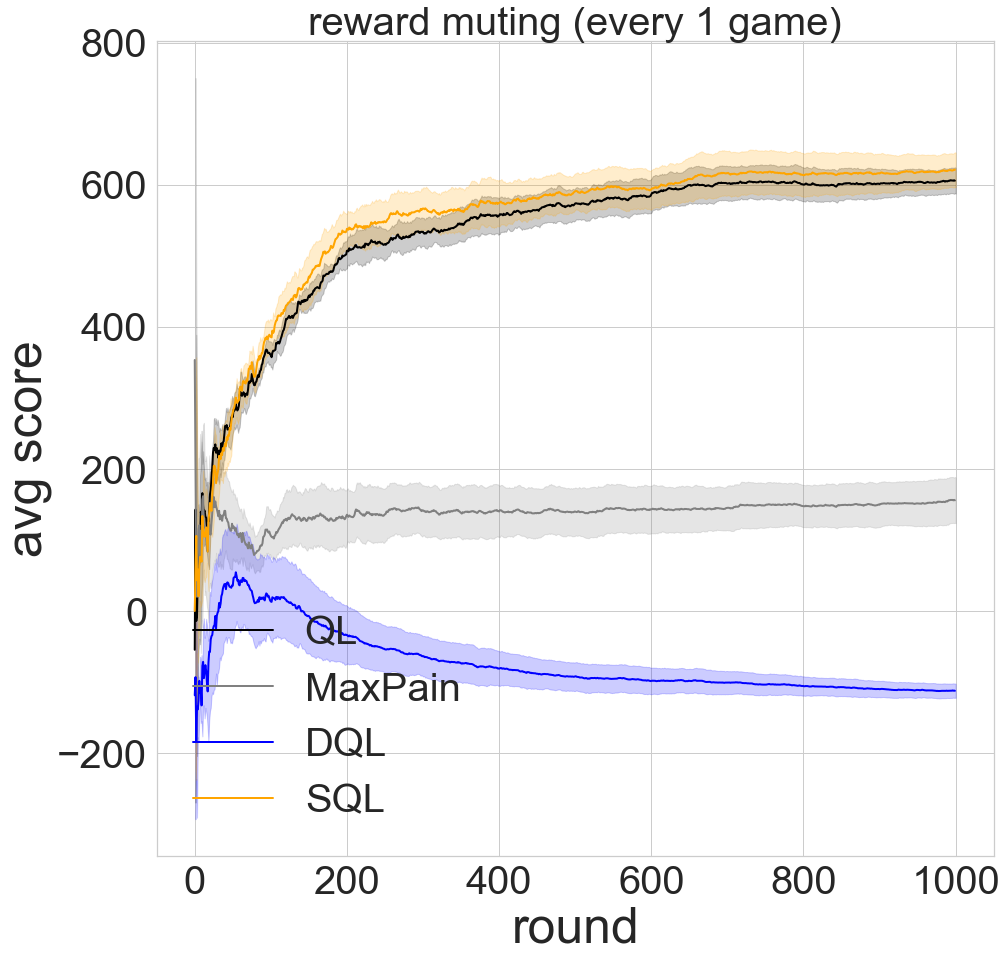}
    \includegraphics[width=0.24\linewidth]{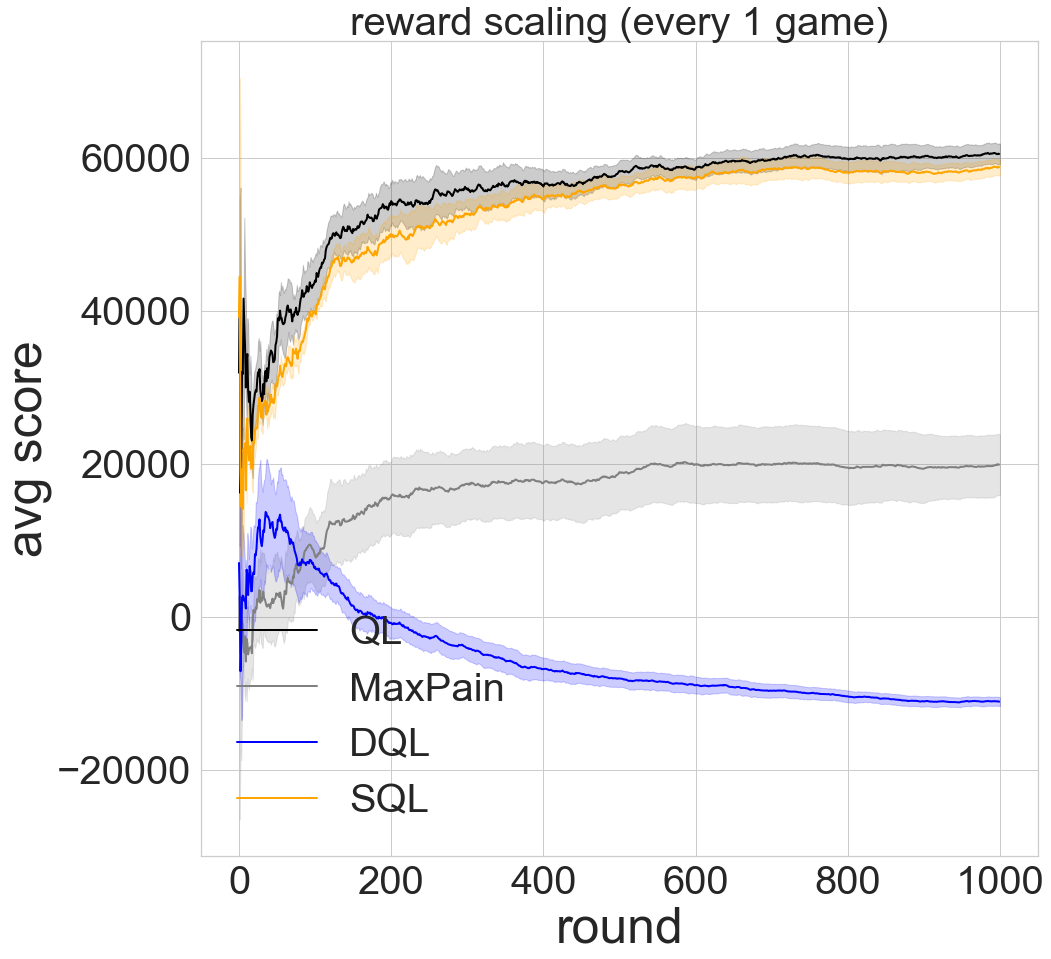}
    \includegraphics[width=0.24\linewidth]{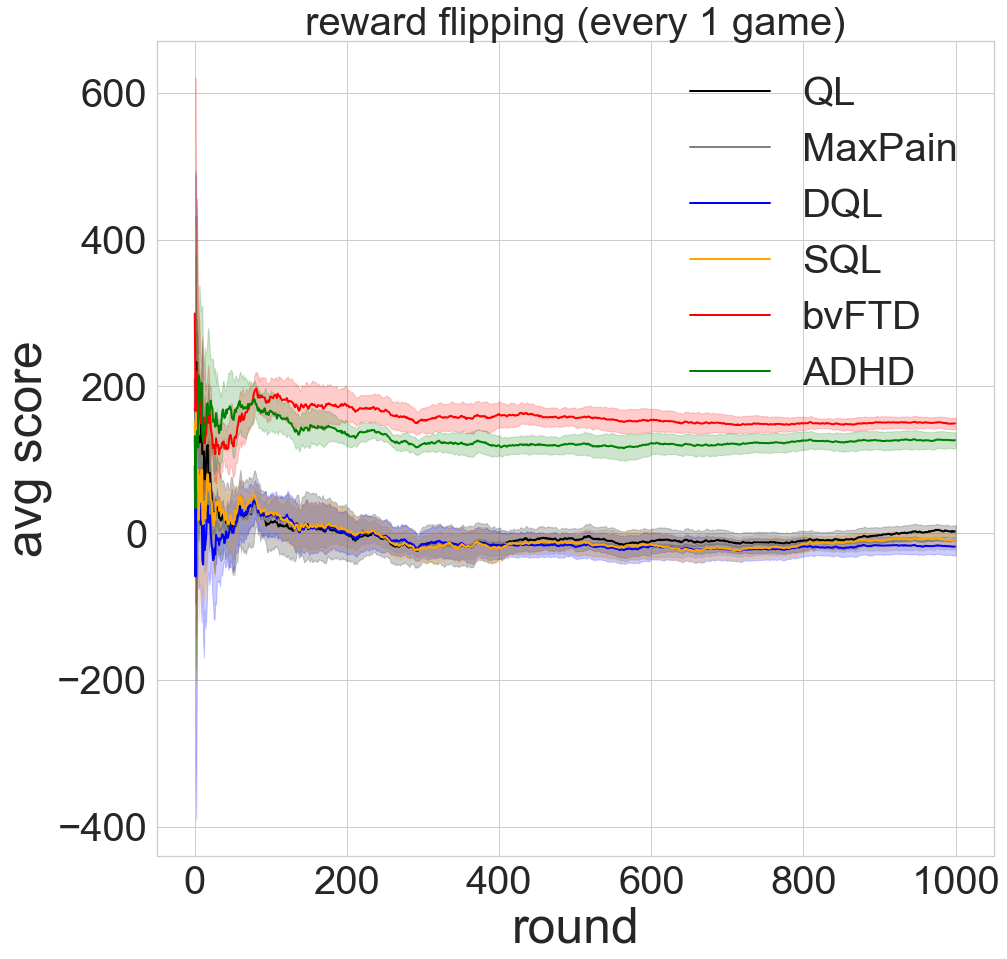}
    \par\caption{\textbf{Average final scores in Pacman game with different stationarities:} (a) stationary; (b) stochastic reward muting by every 1 rounds; (c) stochastic reward scaling by every 1 rounds; (d) stochastic reward flipping by every 1 rounds.}\label{fig:pacman_results_more}
    \vspace{-10em}
\end{figure}

\end{document}